\documentclass{article}

\usepackage{times}
\usepackage{graphicx} %
\usepackage{subfigure} 

\usepackage{natbib}

\usepackage{algorithm}
\usepackage{algorithmic}

\usepackage{hyperref}

\usepackage[accepted]{icml2017}

\graphicspath{ {figs/} }

\usepackage{booktabs}
\usepackage{multirow}
\usepackage{pbox}
\usepackage{amsfonts}
\usepackage{amsmath, amssymb}
\usepackage{placeins}

\usepackage{makecell}

\usepackage[normalem]{ulem}

\newtheorem{definition}{Definition}

\icmltitlerunning{Sharing Residual Units Through Collective Tensor Factorization in Deep Neural Networks}

\begin{document} 

\twocolumn[
\icmltitle{Sharing Residual Units Through Collective Tensor Factorization \\in Deep Neural Networks}

\icmlsetsymbol{equal}{*}

\begin{icmlauthorlist}
\icmlauthor{Chen Yunpeng}{nus}
\icmlauthor{Jin Xiaojie}{nus}
\icmlauthor{Kang Bingyi}{nus}
\icmlauthor{Feng Jiashi}{nus}
\icmlauthor{Yan Shuicheng}{qh,nus}
\end{icmlauthorlist}

\icmlaffiliation{nus}{National University of Singapore }
\icmlaffiliation{qh}{360 AI Institute}

\icmlcorrespondingauthor{Chen Yunpeng}{chenyunpeng@u.nus.edu}

\icmlkeywords{computer vision, pattern recognition, machine learning, ICML}

\vskip 0.3in
]

\printAffiliationsAndNotice{}  %
\setcounter{footnote}{2}
\begin{abstract} 

Residual units are wildly used for alleviating optimization difficulties when building deep neural networks. However, the performance gain does not well compensate the model size increase, indicating low parameter efficiency in these residual units. In this work, we first  revisit the residual function in several variations of residual units and demonstrate that these residual functions can actually be explained with a unified framework based on {generalized block term decomposition}. Then, based on the new explanation, we propose a new architecture, Collective Residual Unit (CRU), which enhances the parameter efficiency of deep neural networks through \emph{collective tensor factorization}. CRU enables knowledge sharing across different residual units using shared factors. Experimental results show that our proposed CRU Network demonstrates outstanding parameter efficiency, achieving comparable classification performance to ResNet-200 with the model size of ResNet-50. By building a deeper network using CRU, we can achieve state-of-the-art single model classification accuracy on ImageNet-1k and Places365-Standard benchmark datasets. (Code and trained models are available on GitHub\footnote{https://github.com/cypw/CRU-Net})

\end{abstract}

\section{Introduction}
\label{l_introduction}
Deep residual networks \citep{he2016deep} are built by stacking multiple \emph{residual units}. Remarkable success has been achieved by deep residual networks for image segmentation \citep{wu2016wider,zhao2016pyramid},  object localization \citep{he2016deep,li2016r}, \emph{etc}. The effectiveness of residual units is attributed to their adopted \emph{identity mapping} and \emph{residual function}. \citet{he2016identity} have explained the importance of the identity mapping in alleviating optimization difficulty. In this work, we focus on the  residual functions. By analyzing various designs of residual functions, we propose a novel architecture with higher parameter efficiency that provides stronger learning capacity.

When the residual network~\citep{he2016deep} was first proposed, the residual function was designed as a three-layer bottleneck architecture consisting of 1$\times$1, 3$\times$3 and 1$\times$1 convolutional filters per layer. The second layer has a less number of channels than the other two convolutional layers. The motivation behind such design is to increase the parameter efficiency by performing the complex 3$\times$3 convolution operations in a lower dimension space.

\begin{figure}[t] 
\begin{center} 
\centerline{\includegraphics[width=\columnwidth]{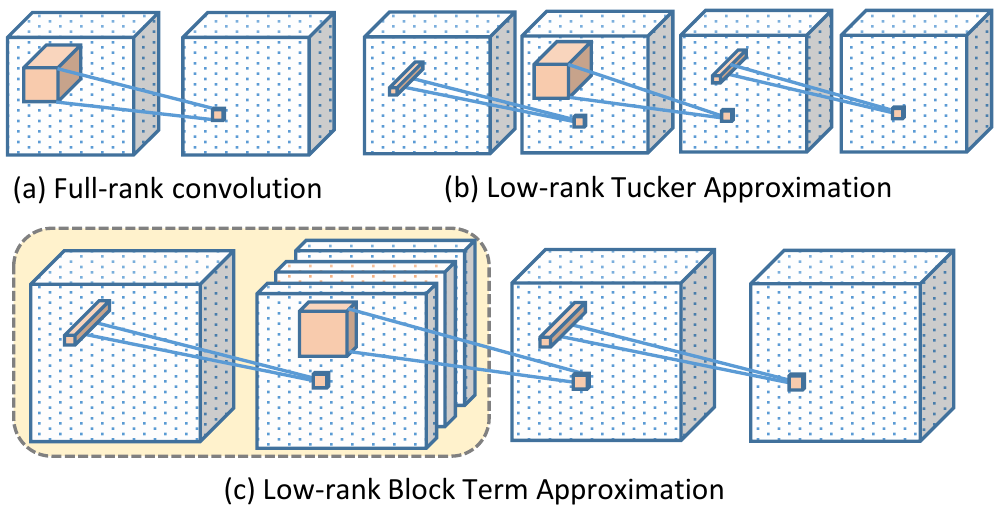}}
\caption{Convolution kernel approximation by different approaches. \textbf{(a)} A full rank convolution layer with a kernel size of $w \times h$. \textbf{(b)} An approximation of (a) by using the low-rank Tucker Decomposition (a special case of Block Term Decomposition when $R=1$). \textbf{(c)} An approximation of (a) by using low-rank Block Term Decomposition. The new proposed CRU shares the first two layers (yellow) across different residual functions.}
\label{first_fig}
\end{center}
\vskip -0.3in
\end{figure}

Since then the residual function has been improved and developed into several different variations. \citet{zagoruyko2016wide} proposed a Wide Residual Network (WDN), which increases the number of channels in the second 3$\times$3 convolutional layer. They found that WDN outperforms the ResNet-152 model with 3 times fewer layers and offers significantly faster speed with roughly the same model size. Recently, \citet{xie2016aggregated} proposed to divide the second 3$\times$3 convolution layer into several groups while keeping the number of parameters almost unchanged. The motivation is to enhance the parameter efficiency by increasing the learning capacity of each bottleneck-shape residual function using transformations aggregated from different paths. Besides, several works~\cite{szegedy2016inception,zhang2016polynet} introduce delicately designed inception architectures into the residual functions and build complex network topology structures with a less number of parameters. However, these inception-style residual functions lack modularity and contain many factors that require expertise knowledge to design.

In this work, we focus on analyzing the various residual functions proposed in ~\citep{he2016deep,he2016identity,zagoruyko2016wide,xie2016aggregated} that are highly modularized and widely used in  different applications. For the first time, our analysis reveals that all of the aforementioned residual functions (that induce different network models) can be unified by viewing them through the lens of tensor analysis~\textemdash~or more concretely a \emph{Generalized Block Term Decomposition} based on the conventional Block Term Decomposition~\cite{lathauwer2008}. With such tensor decomposition, a high order tensor operator (\emph{e.g.},  a set of convolutional kernels operators) is decomposed by a summation of multiple low-rank Tucker operators. Varying the rank of the Tuckers instantiates different residual functions as mentioned above.

Based on this new explanation on residual functions, we further propose a Collective Residual Unit (CRU) architecture %
that enables cross-layer knowledge sharing for different residual units through \emph{collective tensor factorization\footnote{In this work, following the naming conventions in tensor analysis, we interchanged use \emph{factorization} and \emph{decomposition}.}}, illustrated in Figure~\ref{first_fig}. %
With such a novel residual function induced unit, 
information from one residual unit can be reused when building others, leading to significant enhancement of the parameter efficiency in residual networks. We perform extensive experiments on the ImageNet and Place365 datasets to compare the performance of residual networks built upon our proposed CRU and existing residual units. The results clearly verify the outstanding parameter efficiency of our proposed CRU architecture.

The main contributions can be summarized as follows: 

\textbf{1)} We introduce a new perspective for explaining and understanding the  popular convolutional residual networks and unify existing variants of residual functions into a single framework. 

\textbf{2)} Based on the analysis, we propose a novel Collective Residual Unite (CRU) which presents higher  parameter efficiency compared with existing ResNet based models.

\textbf{3)} Our proposed CRU Network achieves  state-of-the-art performance on two large-scale benchmark datasets, This confirms  sharing knowledge across the convolutional layers is  promising  for  pushing the learning capacity and parameter efficiency of state-of-the-art network architectures.

\section{Related Work}
\label{l_related_work}

Tensor decomposition has been introduced in deep learning for a long time, and also the idea of sharing knowledge across different convolutional layers has been proposed ever since the emergence of recurrent neural networks. In this section, we briefly review the related works in both areas and highlight the novelty of this work.

\begin{figure}[t]
\begin{center}
\centerline{\includegraphics[width=\columnwidth]{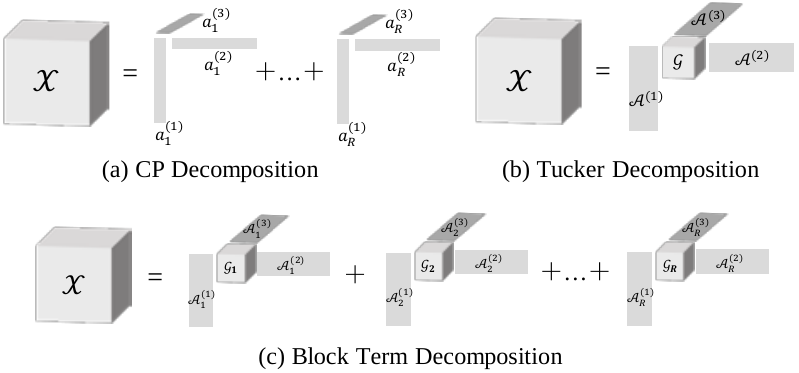}}
\caption{Illustration on factorization of a third order tensor with different tensor decomposition methods.}
\label{tensor_decomp}
\end{center}
\vskip -0.5in
\end{figure}

Mathematically, given a tensor, there are several different ways to factorize it. As can be seen in Figure~\ref{tensor_decomp}, the CANDECOMP/PARAFAC (CP) Decomposition factorizes a tensor as a summation of several tensors with rank equal to one; the Tucker Decomposition factorizes a tensor as a core tensor with multiple 2d matrices. More recently, researchers have combined the CP Decomposition and Tucker Decomposition and proposed a more general decomposition method called \emph{Block Term Decomposition} \citep{lathauwer2008,kolda2009tensor}, where a high-order tensor is approximated in a sum of several low-rank Tuckers. When the rank of each Tucker is equal to one, it degrades to CP Decomposition; when the number of Tuckers equals one, it degrades to Tucker Decomposition. In \citep{novikov2015tensorizing} and \citep{cohen2016deep}, the authors demonstrated that CNNs can be analyzed through tensor factorization, which inspires this work.

One of the many important applications of tensor decomposition is to increase the parameter efficiency. In \citep{lebedev2014speeding}, the authors proposed to compress convolutional layers of a trained model by using CP Decomposition. In \citep{jaderberg2014speeding}, the authors proposed to approximate a initial convolutional kernel tensor by two low-rank components. Similarly, in \citep{garipov2016ultimate}, the authors proposed to decompose fully connected layers by using Tensor-Train Decomposition~\citep{oseledets2011tensor}. These methods either do not support end-to-end training or lack experiments to prove their effectiveness on very large datasets. Sharing knowledge across the neural network is another efficient way to reduce redundancy and increase parameter efficiency. Recurrent Neural Network (RNN) can be seen as a good example where weights are shared across time steps. In \citep{liao2016bridging}, the authors generalized both RNN and ResNet architectures by sharing the entire residual unit throughout the CNNs. In \citep{eigen2013understanding}, the authors proposed to repeat the convolution operation for several times to capture context information. Besides, in \citep{ha2016hypernetworks} the authors proposed to generate the weight for each convolutional layer by using a shallow neural network. However, there is still a gap between the state-of-the-art accuracy and the accuracy of the methods mentioned above, which indicates potential defects within these recurrent architectures and doubts about usefulness of sharing knowledge across layers for the image classification task.

Different from these existing works, we make the first attempt to introduce the Gengeralized Block Term Decomposition into deep learning and derive its corresponding convolutional architectures, which, in turn, forms an unrevealed perspective on various residual functions and further unifies them with a single framework. Based on this new explanation, we propose a novel architecture, Collective Residual Unit, which achieves state-of-the-art accuracy without requiring a pre-trained model and can be easily optimized in an end-to-end manner from scratch. We demonstrate it is an effective way to share information across layers for enhancing parameter efficiency with both mathematical explanation and experimental verification.

\section{Tensor Factorization View on Residual Functions}

In this section, we revisit and explain the existing different residual functions proposed in~\citep{he2016deep,he2016identity,zagoruyko2016wide,xie2016aggregated}, all of which consist of one 1$\times$1, one 3$\times$3 and one 1$\times$1 convolutional layer. 

We start our analysis with introducing the tensor generalized block term decomposition. Then we demonstrate the connection between it and  convolutional architectures. After that, we explain how various residual functions can be unified  into a single framework in the above tensor view.

\subsection{Generalized Block Term Decomposition}

Block Term Decomposition (BTD) was first proposed by \citet{lathauwer2008}, which factorizes a high order tensor into a sum of multiple low-rank Tuckers~\citep{tucker1966some}.
Here, a \emph{tensor} is simply a multidimensional array and the \emph{order} of a tensor refers to the number of dimensions which is also known as \emph{mode}. Throughout the paper, we use calligraphic capital letters, \emph{e.g.} $\mathcal{A} \in \mathbb{R}^{d_1 \times d_2 \times ... \times d_N }$, to denote a tensor, where $d_1, d_2, ...,d_N$ denote the size of each mode. 

Given an $N$-th order tensor $\mathcal{X} \in \mathbb{R}^{d_1 \times d_2 \times ... \times d_N}$, the block term decomposition factorizes it into a sum of $\text{rank-}(d^*_1, d^*_2, ...,d^*_N)$ terms:
\begin{equation} \label{eqn:Def_BTD_org}
\begin{split}
& \mathcal{X} = \sum_{r=1}^{R}{
                      \mathcal{G}_r  
                      \times_1 \mathcal{A}^{(1)}_r 
                      \times_2 \mathcal{A}^{(2)}_r 
                      \times_3 ... 
                      \times_N \mathcal{A}^{(N)}_r} 
~\text{,} \\
&\text{where }\left\{
\begin{aligned}
& ~ \mathcal{G}_r  ~~   \in \mathbb{R}^{d^*_1 \times d^*_2 \times ... \times d^*_N} \\
& \mathcal{A}^{(n)}_r \in \mathbb{R}^{d_n \times d^*_n}, ~ n \in \{1, ..., N\} 
\text{.}
\end{aligned}
\right.\
\end{split}
\end{equation}

Here, $\mathcal{G}$ is known as the core tensor and $\times_n$ denotes the \textit{mode-n} product~\citep{lathauwer2008}. 
Figure~\ref{tensor_decomp}(c) shows an example of applying block term decomposition on a third order tensor.

Elementwisely, the block term decomposition in Eqn.~\eqref{eqn:Def_BTD_org} can be written as 
\begin{equation} \label{eqn:Def_BTD_org_ELE}
\small
\begin{split}
\mathcal{X}(i_1,i_2,...,i_N) & =
	  \sum\limits_{r=1}^{R}
	    \sum\limits_{j_1,...,j_N}
                \mathcal{G}_r(j_1,j_2,...,j_N) \\
                & \mathcal{A}^{(1)}_r(i_1,j_1) \mathcal{A}^{(2)}_r(i_2,j_2) ... \mathcal{A}^{(N)}_r(i_N,j_N)\text{ ,}
\end{split}
\end{equation}
where $i_n \in \{1,...,d_n\}$, $j_n \in \{1,...,d^*_N\}$ and $n \in \{1,...,N\}$.

To analyze multilayer residual functions, one has to take the nolinearity into consideration as a practical deep neural network usually has non-linear activation between two adjacent convolutional layers. Directly applying the conventional block term decomposition to develop tensor representation for the layer-wise computation is non-trivial. Therefore, we follow~\citep{cohen2016convolutional} and propose to generalize the \textit{mode-n} product to functional cases, which in turn generalizes the conventional block term decomposition. 
\begin{definition}[Generalized mode-n product]
	Given an elementwise operation $\sigma$, the generalized model-n product $\times_n^\sigma$, i.e. an operation taking in $\mathcal{G} \in \mathbb{R}^{d^*_1 \times d^*_2 \times ... \times d^*_N}$ and $\mathcal{A} \in \mathbb{R}^{d_n \times d^*_n}$ and returning tensor $\mathcal{G} \times_n^\sigma \mathcal{A} \in \mathbb{R}^{d^*_1 \times ... \times d^*_{n-1} \times d_n \times d^*_{n+1} \times ... \times d^*_N}$, is defined as follows:
\begin{equation}
\label{eqn:Def_GOP}
\begin{split}
(\mathcal{G} \times_n^\sigma \mathcal{A}) = \sigma(\mathcal{G} \times_n \mathcal{A}).
\end{split}
\end{equation}
\end{definition}
By the generalized mode-n product $\times_n^\sigma$, we define the following generalized block term decomposition.
\begin{definition}[Generalized Block Term Decomposition]
	Given an N-th order tensor operator $\mathcal{X}^* \in \mathbb{R}^{d_1 \times d_2 \times ... \times d_N}$, the Block Term Decomposition factorizes it into a sum of $\text{rank-}(d^*_1, d^*_2, ...,d^*_N)$ terms as
\begin{equation}
\label{eqn:Def_BTD_glz}
\begin{split}
& \mathcal{X}^* = \sum_{r=1}^{R}{
                      \mathcal{G}_r  
                      \times_1^\sigma \mathcal{A}^{(1)}_r 
                      \times_2^\sigma \mathcal{A}^{(2)}_r 
                      \times_3^\sigma ... 
                      \times_N^\sigma \mathcal{A}^{(N)}_r} 
~\text{,} \\
&\text{where }\left\{
\begin{aligned}
& ~ \mathcal{G}_r  ~~   \in \mathbb{R}^{d^*_1 \times d^*_2 \times ... \times d^*_N} \\
& \mathcal{A}^{(n)}_r \in \mathbb{R}^{d_n \times d^*_n}, ~ n \in \{1, ..., N\}
\text{.}
\end{aligned}
\right.\
\end{split}
\end{equation}
\end{definition}
Note here the $\mathcal{X}^*$ is no longer a simple multidimensional array but a high order function.

In some cases, one may consider a simplified generalized block term decomposition where specific modes are not factorized. For example, one can keep the first $k$ modes and only factorize the rest modes in $\text{rank-}(\cdot, ..., \cdot, d^*_k, ...,d^*_N)$ as 
\begin{equation}
\label{eqn:Def_BTD_glz_simple}
\begin{split}
& \mathcal{X}^* = \sum_{r=1}^{R}{
                      \mathcal{G}_r  
                      \times_k^\sigma \mathcal{A}^{(k)}_r 
                      \times_{k+1}^\sigma ... 
                      \times_N^\sigma \mathcal{A}^{(N)}_r} 
~\text{,} \\
&\text{where }\left\{
\begin{aligned}
& ~ \mathcal{G}_r  ~~   \in \mathbb{R}^{d_1 \times ... \times d_{k-1} \times d^*_k \times ... \times d^*_N} \\
& \mathcal{A}^{(n)}_r \in \mathbb{R}^{d_n \times d^*_n}, ~ n \in \{k, ..., N\}
\text{.}
\end{aligned}
\right.\
\end{split}
\end{equation}

We now proceed to explain various residual functions through the above introduced  generalized block term decomposition.

\subsection{From GBTD to Convolutional Architectures}

In this subsection we establish connection between the popular convolutional architectures and the Generalized Block Term Decomposition in deep neural networks. Before going into details, we first simplify the Generalized Block Term Decomposition (GBTD) by removing unnecessary decompositions along specific modes in the scenario of residual networks.

For a residual unit, the residual function can be represented as a $4$th order tensor operator $\mathcal{X}^* \in \mathbb{R}^{d_1 \times d_2 \times d_3 \times d_4}$, where $d_1$, $d_2$ represent the width and the height of the filter, and $d_3$, $d_4$ denote the number of the input and output channels, respectively. For most cases, the dimension sizes of $d_1$ and $d_2$ are very low, \textit{e.g.} $d_1=d_2=3$. Therefore, further decomposing it along these two modes is unnecessary. For this reason, we apply the simplified Generalized Block Term Decomposition introduced in Eqn.~\eqref{eqn:Def_BTD_glz_simple} as
\begin{equation} \label{eqn:Def_GBTD_SIMPLE}
\begin{split}
& \mathcal{X}^* = \sum_{r=1}^{R}{\mathcal{G}_r \times_3^\sigma \mathcal{A}^{(3)}_r \times_4^\sigma \mathcal{A}^{(4)}_r} \text{ ,}\\
&\text{where }\left\{
\begin{aligned}
& \mathcal{G}_r \in \mathbb{R}^{d_1 \times d_2 \times d^*_3 \times d^*_4} \\
& \mathcal{A}^{(3)}_r \in \mathbb{R}^{d_3 \times d^*_3} \\
& \mathcal{A}^{(4)}_r \in \mathbb{R}^{d_4 \times d^*_4} 
\text{.}
\end{aligned}
\right.
\end{split}
\end{equation}

Given an input tensor $\mathcal{U} \in \mathbb{R}^{w \times h \times d_3}$, the residual function conducts two-dimensional convolution operation which gives the following output tensor $\mathcal{V} \in \mathbb{R}^{w \times h \times d_4}$:
\begin{equation} \label{eqn:GBT_Cov_org}
\small
\begin{split}
\mathcal{V}(x,y,c) = 
\sum\limits_{r=1}^{R}\sum\limits_{q=1}^{d^*_4}
	\sigma  \bigg[ 
		\sum\limits_{p=1}^{d^*_3}
		\sum\limits_{i=x-\delta_1}^{x+\delta_1}
		\sum\limits_{j=y-\delta_2}^{y+\delta_2}
		\mathcal{G}_r(i-x+\delta, 
\\
		j-y+\delta,p,q)
			\sigma \big( 
			\sum\limits_{m=1}^{d_3}
			\mathcal{U}(i,j,m)
			\mathcal{A}^{(3)}_r(m,q)
			\big)
	\bigg]
	\mathcal{A}^{(4)}_r(c,q)
\end{split}
\end{equation}
where  $\delta_1$ and $\delta_2$ denote  ``half-width'' of the kernel size on each dimension.

If we introduce $\mathcal{T}^{(1)} \in \mathbb{R}^{w \times h \times d^*_3}$ and  $ \mathcal{T}^{(2)} \in \mathbb{R}^{w \times h \times d^*_4}$ to denote intermediate results, Eqn.~\eqref{eqn:GBT_Cov_org} can be simplified as: 
\begin{equation} \label{eqn:eq_GBT_Cov_1}
\begin{split}
\mathcal{T}^{(1)}_r(i,j,q) & \triangleq \sigma \left[ \sum\limits_{m=1}^{d_3}
			         \mathcal{U}(i,j,m)\mathcal{A}^{(3)}_r(m,q) \right],
\end{split}
\end{equation}
\vskip -0.2in
\begin{equation} \label{eqn:eq_GBT_Cov_2}
\begin{split}
\mathcal{T}^{(2)}_r(x,y,q) \triangleq \sigma \bigg[ & \sum\limits_{p=1}^{d^*_3}
			     \sum\limits_{i=x-\delta_1}^{x+\delta_1}
			       \sum\limits_{j=y-\delta_2}^{y+\delta_2} 
			       \mathcal{T}^{(1)}_r(i,j,p) 
\\
&			       
			          \mathcal{G}_r(i-x+\delta,j-y+\delta,p,q) \bigg],
\end{split}
\end{equation}
\vskip -0.3in
\begin{equation} \label{eqn:eq_GBT_Cov_3}
\begin{split}
\mathcal{V}(x,y,c) & = \sum\limits_{r=1}^{R}
			   \sum\limits_{q=1}^{d^*_4}
			     \mathcal{T}^{(2)}_r(x,y,q)
			     \mathcal{A}^{(4)}_r(c,q).
\end{split}
\end{equation}
\vskip -0.2in

Figure~\ref{first_fig} (c) shows the corresponding convolutional architectures of Eqns.~\eqref{eqn:eq_GBT_Cov_1} \eqref{eqn:eq_GBT_Cov_2} \eqref{eqn:eq_GBT_Cov_3}. Specifically, the input tensor is first convoluted by a 1$\times$1 convolution kernel and then passed to a group convolutional layer, which equally separates the input tensors into $R$ groups along the third mode and conducts convolution operation within each group separately. After that, the results are mapped by 1$\times$1 convolutional kernel and aggregated as the final result. 

To the best of our knowledge, this is the first work to introduce and generalize the conventional Block Term Decomposition for analyzing convolutional neural network.

\subsection{Unifying Residual Functions} 

The convolutional architecture that we have derived in Figure~\ref{first_fig} shows a strong relation with various residual functions. In this subsection, we give a comprehensive analysis on these different residual functions and explain them within a single framework based on the introduced Generalized Block Tensor Decomposition.

The conventional residual functions proposed in \cite{he2016deep,he2016identity,zagoruyko2016wide} are special cases of the Generalized Block Tensor Decomposition when $R=1$, $d_3 = d_4 = \texttt{width of the shortcut} $ and $d^*_3 = d^*_4 =  \texttt{width of the bottleneck}$. Figure~\ref{first_fig}(b) demonstrates the case when $R$ is set to 1 which is in the exactly same form of vanilla residual function~\citep{he2016deep,he2016identity} and the wide residual network~\cite{zagoruyko2016wide}. Specifically, the tensor $\mathcal{A}^{(3)}$ in Eqn.~\eqref{eqn:Def_GBTD_SIMPLE} corresponds to the first 1$\times$1 convolutional kernel tensor where the number of input channels is $d_3$ and the number of output channels is $d^*_3$. Similarly, the tensor $\mathcal{A}^{(4)}$ corresponds to the last 1$\times$1 convolutional kernel tensor and $\mathcal{G}$ corresponds to the second 3$\times$3 convolutional kernel tensor. The difference between the standard residual unit and the wide residual unit is that the latter uses a core kernel tensor with higher rank.

Note that most residual functions add a batch normalization layer before (or after) the convolutional layer to avoid the covariance shift problem~\citep{ioffe2015batch}. However, adding the batch normalization layer does not affect our derived results, since this operation is mathematically an elementwise linear function which can be absorbed into the nearest convolutional kernel tensor.

Interestingly, Eqn.~\eqref{eqn:Def_GBTD_SIMPLE} can be seen as an aggregation of multiple transformations, which has the same form as Eqn. (2) proposed in a recently published paper~\cite{xie2016aggregated}. The \textit{cardinality} proposed in \cite{xie2016aggregated} directly corresponds to $R$ in Eqn.~\eqref{eqn:Def_GBTD_SIMPLE}, which refers to the number of low-rank Tuckers. The $r$-th low-rank Tucker, \textit{i.e.} $\mathcal{G}_r \times_3 \mathcal{A}^{(3)}_r \times_4 \mathcal{A}^{(4)}_r$, corresponds to the $r$-th kernel tensor of the transform function $\mathcal{T}_r(\mathcal{U})$. 
Here, the $\mathcal{U}$ is the input to the residual function.

In other words, the residual function proposed in ResNeXt is a special case of the Generalized Block Term Decomposition with the settings below:
\begin{equation} \label{eq_Tucker_setting_1}
\begin{split}
\left\{
\begin{aligned}
R &=  \mathrm{ cardinality } \\
d_3 &= d_4  =  \mathrm{ width~ of ~the~ shortcut}\\
d^*_3 &= d^*_4  =  \frac{\mathrm{ width~ of ~the ~bottleneck}}{R},
\end{aligned}
\right.
\end{split}
\end{equation}
where the \emph{width of the bottleneck} simply refers to the number of channels for the second 3$\times$3 grouped convolutional layer.

Such observation indicates that their new proposed cardinality is essentially the number of low rank Tuckers. When the number of parameters is fixed, the higher $R$ becomes, the lower the representation ability of each Tucker will be, indicating that $R$ may not be proportional with the learning capacity. We have verified this in our experiments by setting the $\texttt{ width of the bottleneck}= \texttt{width of the bottleneck}$ for each 3$\times$3 convolutional layer within the residual function.

\section{Collective Residual Unit Networks}
In this section, we introduce a novel residual unit -- Collective Residual Unit, based on a collective tensor factorization method that is specifically designed for sharing information across residual units. Below, we first describe the new  collective tensor factorization method and then explain the proposed CRU Network, followed by its complexity analysis.

\subsection{Collective Tensor Factorization}
In highly modularized deep residual networks, residual units with the same architecture are stacked together. Removing any one of the residual units will not result in obvious performance drop~\citep{veit2016residual}, which indicates great redundancy across residual units. Motivated by this observation, we propose to reduce the redundancy by reusing (sharing) information from one residual function  for constructing another. We achieve this goal by simultaneously factorizing multiple convolutional kernel tensor operators and sharing factors across them. We refer to this approach as \emph{collective tensor factorization} whose details are given below.%

In particular, for a highly modularized residual network with $L$ similar residual functions stacked together, we concatenate each convolutional kernel tensor $\mathcal{X}^*_l \in R^{d_1 \times d_2 \times d_3 \times d_4}$ along its $4$th mode as $\mathcal{X}^+ = \left[\mathcal{X}^*_1, \mathcal{X}^*_2, ..., \mathcal{X}^*_L\right]$, where $\mathcal{X}^+ \in R^{d_1 \times d_2 \times d_3 \times (L*d_4) }$. 	
Then factorize $\mathcal{X}^+$ by using the \emph{Generalized Block Term Decomposition} with $\text{rank-}(\cdot,\cdot,d^*_3,d^*_4)$, as shown in Eqn.~\eqref{eq_share_1}.
\begin{equation} \label{eq_share_1}
\begin{split}
& \mathcal{X}^+ = \sum_{r=1}^{R}{\mathcal{G}_r \times_3^\sigma \mathcal{A}^{(3)}_r \times_4^\sigma \mathcal{A}^{(4)}_r} \text{ ,}\\
&\text{where, }\left\{
\begin{aligned}
& \mathcal{G}_r \in \mathbb{R}^{d_1 \times d_2 \times d^*_3 \times d^*_4} \\
& \mathcal{A}^{(3)}_r \in \mathbb{R}^{d_3 \times d^*_3} \\
& \mathcal{A}^{(4)}_r \in \mathbb{R}^{(L*d_4) \times d^*_4} 
\text{.}
\end{aligned}
\right.
\end{split}
\end{equation}

This is equivalent to decomposing each $\mathcal{X}^*_l$ with $\text{rank-}(\cdot,\cdot,d^*_3,d^*_4)$ separately and then sharing the first two factor terms. 

\begin{figure}[t] 
\begin{center}
\centerline{\includegraphics[width=\columnwidth]{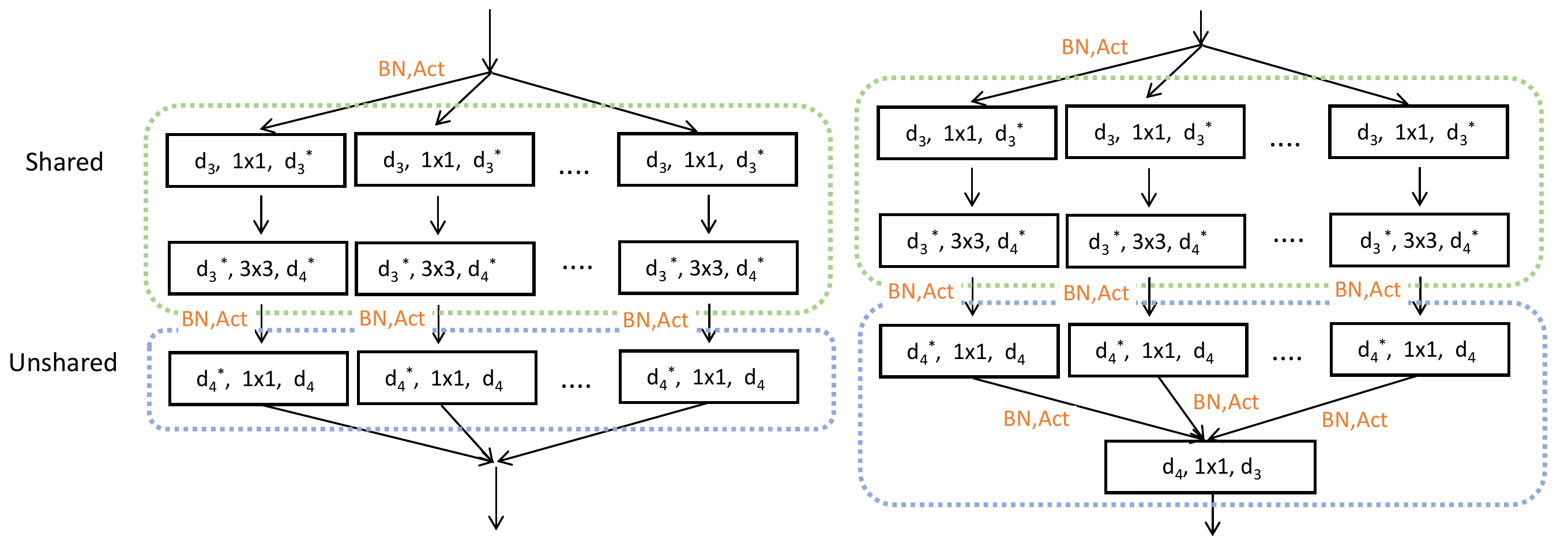}}
\caption{The proposed collective residual unit architecture based on collective tensor factorization. \textbf{Left} is the standard form without consider the nonlinearity; \textbf{Right} is the improved form with better nonlinearity. ``BN'' refers to ``Batch Normalization'', ``Act'' refers to ``Activation''. The parameters of the first two layers (highlighted by the green bounding box) are shared \emph{across different residual units}, while the other layer(s) are not shared. Best viewed in zoomed PDF.}
\label{fig_new_block}
\end{center}
\vskip -0.3in
\end{figure}

By the collective tensor factorization above, parameters are shared across different residual units. During the learning stage, different from the unshared version, the gradient information from each kernel would aggregate together before updating the shared factors, making the learning process more efficient. Figure~\ref{fig_new_block} (left) shows the corresponding convolution structure, where the first 1$\times$1 convolutional layer and the second 3$\times$3 convolutional layer are shared across $L$ different layers, and at the same time each layer has its own $\mathcal{A}^{(4)}_r$ which is not shared with other layers. 

We name this new residual unit as \emph{Collective Residual Unit (CRU)} and build the new CRU Network by stacking multiple CRUs.
 
\subsection{Proposed Network Architecture}
 
Our proposed CRU Network is built by stacking multiple modularized Collective Residual Units. Since the most recently proposed ResNeXt~\cite{xie2016aggregated} is a special case of our proposed CRU when the parameters are not shared across units, we simply adopt general settings in \cite{xie2016aggregated} and do ablation experiments by replacing the residual units with our new proposed CRUs. We keep the model size roughly unchanged (or even a little smaller) compared with the vanilla residual network~\citep{he2016deep} throughout our design, to ensure that the improvement comes from the higher parameter efficiency.

\subsubsection{Activation and batch normalization}
The analysis in the above ignored the batch normalization layers. Here we consider adding them back to 
build a complete deep neural network. 

When information is not shared across the layers, which means no parameter sharing, the batch normalization layers can be directly added back. However, when information is shared across layers, adding them back becomes complicated since the batch normalization~\citep{ioffe2015batch} 
cannot be shared across layers by its nature. Moreover, since the activation function is usually combined with the batch normalization, it would lose one ``BN,Act'' as shown in Figure~\ref{fig_new_block} (left), which would lead to lack of nonlinearty and thus affect the final performance.

{To solve this problem, we propose to append another 1$\times$1 convolution layer after the second 1$\times$1 convolutional layer as shown in Figure~\ref{fig_new_block} (right). Note that introducing this 1$\times$1 convolutional layer does not increase the overall model size compared with existing models, benefiting from parameter sharing by CRUs.} Also, because the 1$\times$1 convoltion operation has very low computational cost, it would not greatly affect the computational efficiency. 
\subsubsection{Overall Setting}

Table~\ref{tab_detail_settings} presents the detailed overall setting of our proposed method and several baseline methods. The CRU Network, noted as CRU-Net, is designed by stacking multiple similar CRUs. %
We set the number of output channels the same for the first three convolutional layers. To avoid potential
optimization difficulties, parameters are only shared within every six layers, and for those with less than six layers, we do not share parameters across units so that the overall setting is consistent. 

The notation ``\texttt{$R \times d_3^*$d}'' is introduced to represent the network settings, where $R$ is the number of low-rank Tuckers which corresponds to the number of groups, $d_3^*=d_4^*$ controls the rank of each Tucker, see Eqn.~\eqref{eqn:Def_GBTD_SIMPLE}. For CRU-Net, we use ``@stage'' to indicate stage where we adopt CRU. For example, in Table~\ref{tab_detail_settings}, ``$32\times4$d @$\times$28$\times$14'' denotes the $R=32$, $d^*_3=d^*_4=4$, adopt CRU at \emph{conv3} and \emph{conv4}.

The width of the CRU Network is computed to make the overall model size roughly the same as the vanilla ResNet. The number of groups is simply set to its maximum number, \emph{i.e.} equal to the channels size. One can set a different number of groups in order to find the optimal setting. However, since the NVIDIA CuDNN library does not support group convolution yet, making the grouped convolution operation slow in practical implementation and thus the number of groups is almost impossible to tune, we simply set it equal to the number of channels.

\begin{table*}[t]
\vskip -0.1in
  \tiny
  \centering
  \setlength\tabcolsep{3pt}
  \caption{Comparison of our proposed CRU Network (CRU-Net) and different residual networks. We compare our proposed CRU-Net with three baseline methods: vanilla ResNet~\citep{he2016deep}, ResNeXt~\citep{xie2016aggregated}, and ResNeXt with the highest $R$ for each 3$\times$3 convolutional layer in the residual unit. We also show the detailed setting of a deeper CRU-Net (CRU-Net-116) with a slightly less number of parameters than the vanilla ResNet-101~\citep{he2016deep} (\#params: $44.31 \times 10^6$). }
    \begin{tabular}{c|c|c|c|c|c|c}
    \toprule
    stage & output 
    & ResNet-50 
    & ResNeXt-50 (32$\times$4d) 
    & ResNeXt-50 (N$\times$1d) 
    & \textbf{CRU-Net-56  (32$\times$4d @$\times$14)}
    & \textbf{CRU-Net-116 (32$\times$4d @$\times$28$\times$14)} \\
    \midrule
    conv1 & 112x112
    & $7 \times 7$, 64, stride 2 
    & $7 \times 7$, 64, stride 2 
    & $7 \times 7$, 64, stride 2 
    & $7 \times 7$, 64, stride 2 
    & $7 \times 7$, 64, stride 2 \\
    \midrule
    \multirow{2}[4]{*}{conv2} & \multirow{2}[4]{*}{56x56} 
    & $3 \times 3$ max pool, stride 2 
    & $3 \times 3$ max pool, stride 2 
    & $3 \times 3$ max pool, stride 2 
    & $3 \times 3$ max pool, stride 2 
    & $3 \times 3$ max pool, stride 2 \\
	\cmidrule{3-7}          &      
	&  $\left[\begin{array}{l} \textsc{1$\times$1, 64}   \\ \textsc{3$\times$3,  64, R=1 }  \\ \textsc{1$\times$1, 256}  \end{array} \right] \times 3 $
	&  $\left[\begin{array}{l} \textsc{1$\times$1, 128}  \\ \textsc{3$\times$3, 128, R=32}  \\ \textsc{1$\times$1, 256}  \end{array} \right] \times 3 $
	&  $\left[\begin{array}{l} \textsc{1$\times$1, 136}  \\ \textsc{3$\times$3, 136, R=136} \\ \textsc{1$\times$1, 256}  \end{array} \right] \times 3 $
	&  $\left[\begin{array}{l} \textsc{1$\times$1, 128}  \\ \textsc{3$\times$3, 128, R=32}  \\ \textsc{1$\times$1, 256}  \end{array} \right] \times 3 $
	&  $\left[\begin{array}{l} \textsc{1$\times$1, 128}  \\ \textsc{3$\times$3, 128, R=32}  \\ \textsc{1$\times$1, 256}  \end{array} \right] \times 3 $ \\
    \midrule
    conv3 & 28$\times$28
	&  $\left[\begin{array}{l} \textsc{1$\times$1, 128}  \\ \textsc{3$\times$3, 128, R=1 }  \\ \textsc{1$\times$1, 512}  \end{array} \right] \times 4 $
	&  $\left[\begin{array}{l} \textsc{1$\times$1, 256}  \\ \textsc{3$\times$3, 256, R=32}  \\ \textsc{1$\times$1, 512}  \end{array} \right] \times 4 $
	&  $\left[\begin{array}{l} \textsc{1$\times$1, 272}  \\ \textsc{3$\times$3, 272, R=272} \\ \textsc{1$\times$1, 512}  \end{array} \right] \times 4 $
	&  $\left[\begin{array}{l} \textsc{1$\times$1, 256}  \\ \textsc{3$\times$3, 256, R=32}  \\ \textsc{1$\times$1, 512}  \end{array} \right] \times 4 $
	&  $\left[\begin{array}{l} \textbf{1$\times$1, 352}  \\ \textbf{3$\times$3, 352, R=352} \\ \textsc{1$\times$1, 352} \\ \textsc{1$\times$1, 512}  \end{array} \right] \times 6 $ \\
    \midrule
    conv4 & 14$\times$14
	&  $\left[\begin{array}{l} \textsc{1$\times$1, 256}  \\ \textsc{3$\times$3, 256, R=1  } \\ \textsc{1$\times$1, 1024} \end{array} \right] \times 6 $
	&  $\left[\begin{array}{l} \textsc{1$\times$1, 512}  \\ \textsc{3$\times$3, 512, R=32}  \\ \textsc{1$\times$1, 1024} \end{array} \right] \times 6 $
	&  $\left[\begin{array}{l} \textsc{1$\times$1, 544}  \\ \textsc{3$\times$3, 544, R=544} \\ \textsc{1$\times$1, 1024} \end{array} \right] \times 6 $
	&  $\left[\begin{array}{l} \textbf{1$\times$1, 640}  \\ \textbf{3$\times$3, 640, R=640} \\ \textsc{1$\times$1, 640} \\ \textsc{1$\times$1, 1024} \end{array} \right] \times 6 $
	&  $\left[\left[\begin{array}{l} \textbf{1$\times$1, 704}  \\ \textbf{3$\times$3, 704, R=704}  \\ \textsc{1$\times$1, 704} \\ \textsc{1$\times$1, 1024} \end{array} \right] \times 6 \right] \times 3 $ \\
    \midrule
    conv5 & 7$\times$7
	&  $\left[\begin{array}{l} \textsc{1$\times$1, 512}  \\ \textsc{3$\times$3,  512, R=1 }   \\ \textsc{1$\times$1, 2048} \end{array} \right] \times 3 $
	&  $\left[\begin{array}{l} \textsc{1$\times$1, 1024} \\ \textsc{3$\times$3, 1024, R=32}   \\ \textsc{1$\times$1, 2048} \end{array} \right] \times 3 $
	&  $\left[\begin{array}{l} \textsc{1$\times$1, 1088} \\ \textsc{3$\times$3, 1088, R=1088} \\ \textsc{1$\times$1, 2048} \end{array} \right] \times 3 $
	&  $\left[\begin{array}{l} \textsc{1$\times$1, 1024} \\ \textsc{3$\times$3, 1024, R=32}   \\ \textsc{1$\times$1, 2048} \end{array} \right] \times 3 $
	&  $\left[\begin{array}{l} \textsc{1$\times$1, 1024} \\ \textsc{3$\times$3, 1024, R=32}   \\ \textsc{1$\times$1, 2048} \end{array} \right] \times 3 $ \\
    \midrule
          & 1$\times$1
    & \pbox{20cm}{global average pool \\ 1000-d fc, softmax} 
    & \pbox{20cm}{global average pool \\ 1000-d fc, softmax} 
    & \pbox{20cm}{global average pool \\ 1000-d fc, softmax} 
    & \pbox{20cm}{global average pool \\ 1000-d fc, softmax} 
    & \pbox{20cm}{global average pool \\ 1000-d fc, softmax} \\
    \midrule
    \multicolumn{2}{c|}{\# params} 
    & $\mathbf{25.5} \times 10^6$ 
    & $\mathbf{25.0} \times 10^6$ 
    & $\mathbf{24.9} \times 10^6$ 
    & $\mathbf{25.5} \times 10^6$ 
    & $\mathbf{43.7} \times 10^6$ \\
    \midrule
    \multicolumn{2}{c|}{FLOPs} 
    & $\mathbf{4.1} \times 10^9$ 
    & $\mathbf{4.2} \times 10^9$ 
    & $\mathbf{4.3} \times 10^9$ 
    & $\mathbf{4.9} \times 10^9$ 
    & $\mathbf{13.3} \times 10^9$  \\
    \bottomrule
    \end{tabular}%
  \label{tab_detail_settings}%
\vskip -0.1in
\end{table*}%

\subsection{Model Complexity}

Here, we follow \citep{he2016deep} and \citep{xie2016aggregated} to evaluate the model complexity from two aspects: the overall model size and the computational cost.

\textbf{Model Size}
We compute the model size by counting the number of trainable parameters within the model. The width of the bottleneck is adjusted to make the overall model size roughly the same or even less than the baseline model. Table~\ref{tab_detail_settings} shows the model size for each model.

\textbf{Computational Cost}
The theoretical computational cost is shown in Table~\ref{tab_detail_settings}. However, in practice, the time cost would be much higher since the NVIDIA CuDNN library does not yet support the convolutional layer with $\texttt{number of group} > 1$. As a result, many deep learning frameworks, \emph{e.g.} MXNet~\citep{chen2015mxnet}, use a sequential way to process each group separately. When the computational loads are not enough, the communication consumption and task management cost would dominate and significantly slow down the speed. 

\section{Implementation}

\noindent
We implement our proposed method by using MXNet \citep{chen2015mxnet} on a cluster with 68 GPUs. We summarize our implementation details as follows.

\textbf{Data Augmentation} 
We adopt both color and spatial augmentations on pre-shuffled input raw images. Specifically, the whole image is first added a random noise vector in HSL color space and then randomly cropped with an area 
ranging from 8\% to 100\% of the whole image with aspect ratio ranging from $3/4$ to $4/3$. After that, the cropped image is randomly horizontally flipped and resized to 224$\times$224 before fed into the network. The random 
noise vector is sampled from $H (hue) \in [-20,20]$, $S (saturation) \in [-40,40]$, and $L (lightness) \in [-50,50]$.

The differences between our data augmentation and that in \citep{xie2016aggregated} are two-fold: \textbf{(a)} We conduct the color augmentation in HSL color space instead of HSV color space. \textbf{(b)} We do not use the PCA lighting \citep{krizhevsky2009learning}, which is orthogonal to our augmentations thus may further improve the performance of our method.

\textbf{Training Setting} 
Training CNNs on distributed GPUs can be much more difficult than training on a single node, since the batch size can be significantly larger and the number of iterations will thus be insufficient. To achieve the best performance, the weight decay and the base learning rate are set to $0.0005$ and $0.1$ for all CNNs with 50 layers, while they are set to $0.0002$ and $\sqrt{0.1}$ for all deeper CNNs. The learning rate is decreased by a factor $0.1$ when the validation accuracy gets saturated. Throughout the experiments, we use SGD with nesterov~\citep{kingma2014adam} with a mini-batch size of $32$ for each GPU and update in a synchronized manner.

\textbf{Testing Setting} 
Without specific notification, we evaluate the classification error rate on single 224$\times$224 center crop from the raw input image with short length equal to 256, which is the same setting as \citep{he2016deep,he2016identity,xie2016aggregated}.

\section{Experiments}

We evaluate our proposed method on two widely used large-scale datasets, the ILSVRC-1000 classification dataset \citep{ILSVRC15}, a.k.a ImageNet-1k, and the Places365-Standard dataset \citep{zhou2016places}, a.k.a Places365-Standard.  For the ImageNet-1k dataset, we report single crop validation error rate following \citep{he2016deep}. For Places365-Standard, we report 10 crops validation accuracy following \citep{zhou2016places}. Throughout the experiments, we use ``(ours)'' to denote the baseline method that is reproduced by ourselves.

\subsection{Results on ImageNet-1k}

The ImageNet-1k dataset is for object classification, with about 1.28 million high-resolution images of 1,000 categories. For this dataset, we first conduct ablation experiments to study the properties of our proposed method. Then we compare our proposed model with state-of-the-art models by building deeper and wider CRU networks.

Firstly, we conduct a set of experiments to study the effect of the number of Tuckers, \emph{i.e.} $R$, on the performance. Here, we fix the model size roughly the same and vary the $R$, as shown in Table~\ref{l_tab_imnet50}. Since the overall number of parameters is constrained, a bigger number of Tuckers means the lower representation ability for each Tucker (see Eqn.~\eqref{eqn:Def_GBTD_SIMPLE}). As can be seen in the last row of Table~\ref{l_tab_imnet50}, when we set the number of $R$ to its maximum value for each residual function, the error rate increased from $22.1\%$ to $22.5\%$. This might be caused by the insufficient learning capacity of each Tucker. The best setting, as can be seen in the first three rows, is to double the rank of each Tucker when the input increases. Such observation indicates that the number of groups, a.k.a \emph{Cardinality}~\citep{xie2016aggregated}, is not proportional with the performance. 

Secondly, we evaluate our proposed network comparing with other state-of-the-art residual networks. Table~\ref{l_tab_imagenet1k_50ly_all} summarizes different variations of residual network under the same model size. The ResNet-200 stands for the best published performance reported in \citep{he2016identity}. As can be seen from this table, our proposed model under the setting of ``32x4d'' achieves $21.9\%$ top1 error rate comparing with $22.2\%$ for ResNeXt. When we change the setting from ``32x4d'' to ``136x1d'', the performance of ResNeXt seen to saturate. While our proposed network achieve $21.7\%$ top-1 error rate which is comparable with ResNet-200~\citep{he2016identity}.

\begin{table}[t]
\caption{Single crop validation error rate of residual networks with different $R$ on ImageNet-1k dataset.}
\label{l_tab_imnet50}
\begin{center}
\renewcommand{\arraystretch}{1.3}
\resizebox{0.49\textwidth}{!}{
\begin{small}
\begin{tabular}{lcccr}
\hline
\abovespace\belowspace
Method    			&  setting 	& model size & top-1 err.(\%) \\
\hline
ResNeXt-50 (ours)   	&  2 x 40d	&   98 MB  	 & 22.8  \\
ResNeXt-50 (ours)   	&  32 x 4d	&   96 MB	 & 22.2  \\
ResNeXt-50 (ours)   	&  136 x 1d	&   97 MB 	 & 22.1  \\
ResNeXt-50 (ours)   	&  N x 1d	&   96 MB	 & 22.5  \\
\hline
\end{tabular}
\end{small}
}
\end{center}
\vskip -0.2in
\end{table}

Finally, we increase the model size of our proposed model to build a more complex CRU Network and compare its performance with the state-of-the-art residual networks. As can be seen from Table~\ref{l_tab_imagenet1k_200ly_all}, our proposed model achieves the best performance. However, the improvement is quite marginal. We believe the very deep CRU-Net's learning capacity is more than enough for handling this dataset, since severe over-fitting problem is observed when doubling the model size of CRU-Net-116. It indicates that the CRU-Net-116 or even CRU-Net-56 is enough for handling this dataset.

\begin{table}[t]
  \caption{Single crop validation error rate on ImageNet-1k dataset. }
  \label{l_tab_imagenet1k_50ly_all}
  \renewcommand{\arraystretch}{1.3}  
  \resizebox{0.49\textwidth}{!}{
  \begin{small}
  \begin{tabular}{lcccr}
	\hline
	\abovespace\belowspace
	Method    							&  setting 	& model size & top-1 err.(\%) \\
	\hline
	ResNet-50~\cite{he2016deep}			&  1 x 64d	&   98 MB 	 & 23.9  		        \\
	ResNet-200~\cite{he2016identity} 	&  1 x 64d	&  247 MB	 & \textbf{21.7}    \\
	\hline
	ResNeXt-50~\cite{xie2016aggregated}	&  2 x 40d	&   98 MB 	 & 23.0    \\
	ResNeXt-50~\cite{xie2016aggregated}	&  32 x 4d	&   96 MB	 & 22.2    \\
	\hline
	ResNeXt-50 (ours)   					&  2 x 40d	&   98 MB  	 & 22.8    \\
	ResNeXt-50 (ours)   					&  32 x 4d	&   96 MB	 & 22.2    \\
	ResNeXt-50 (ours)   					& 136 x 1d	&   97 MB 	 & 22.1    \\
	\hline
	CRU-Net-56 @$\times$14				&  32 x 4d 	&   98 MB 	 & 21.9    \\
	CRU-Net-56 @$\times$14				& 136 x 1d	&   98 MB 	 & \textbf{21.7}    \\
	\hline
  \end{tabular}
  \end{small}
  }
\vskip -0.1in
\end{table}

\begin{table}[t]
  \caption{Comparison with state-of-the-art residual networks. Single crop validation error rate on ImageNet-1k dataset.}
  \label{l_tab_imagenet1k_200ly_all}
  \renewcommand{\arraystretch}{1.3}  
  \resizebox{0.49\textwidth}{!}{
  \begin{small}
  \begin{tabular}{lcccr}
	\hline
	\abovespace\belowspace
	Method    		&  setting 	& model size & \makecell[c]{top-1 \\ err.(\%)} & \makecell[c]{top-5 \\ err.(\%)} \\
	\hline
	ResNet-101~\cite{he2016deep}	    			&  1 x 64d	&  170 MB 	 & 22.0    &   6.0    \\
	ResNeXt-101~\cite{xie2016aggregated}		&  32 x 4d	&  170 MB	 & 21.2    &   5.6    \\
	CRU-Net-116 @$\times$28$\times$14 		&  32 x 4d	&  168 MB	 & 20.6    &   5.4    \\
	\hline
	ResNet-200~\cite{he2016deep}				&  1 x 64d	&  247 MB 	 & 23.0    &   5.8	  \\
	WRN~\cite{zagoruyko2016wide}				&  1 x 128d	&  263 MB  	 & 21.9    &   5.8   \\
	ResNeXt-101, wider ~\cite{xie2016aggregated} &  64 x 4d	&  320 MB	 & 20.4    &   5.3    \\
	ResNeXt-101, wider (ours)					&  64 x 4d	&  320 MB	 & 20.4    &   5.3    \\
	CRU-Net-116, wider @$\times$28$\times$14 &  64 x 4d	&  318 MB	 & \textbf{20.3}  & \textbf{5.3}    \\
	\hline
  \end{tabular}
  \end{small}
  }
\vskip -0.2in
\end{table}

\subsection{Results on Places365-Standard} 

We further evaluate our proposed network on one of the largest scene classification datasets -- Places365-Standard~\citep{zhou2016places}. The Places365-Standard consists of 1.8 million images of 365 scene categories. Different from the object classification task where most distinguishable parts determine the image label, the scene recognition requires a more logical reasoning ability and a larger receptive filed.

Table~\ref{l_tab_places365} shows the results of different models on Places365-Standard dataset. The results in first four rows are provided by \cite{zhou2016places} and the last two rows are from our implementation. Our proposed method achieves the best classification accuracy compared with other methods. Comparing with the vanilla ResNet-152, our proposed method improves the top-1 performance by absolute value of $1.2\%$ with significantly smaller model size ($163$MB \emph{v.s.} $226$MB), which again confirms the effectiveness of our proposed CRU Network.

\begin{table}[t]
\caption{Single model, 10 crops validation accuracy on Places365-Standard dataset. Results in the first four rows are from \citep{zhou2016places}. Results in the last two rows are from our implementation.}
\label{l_tab_places365}
\renewcommand{\arraystretch}{1.3}
\resizebox{0.49\textwidth}{!}{
\begin{small}
\begin{tabular}{lcccr}
\hline
Method    			& setting    & model size &  \makecell[c]{top-1 \\ acc.(\%)} & \makecell[c]{top-5 \\ acc.(\%)} \\
\hline
AlexNet    							&  	--   	 	 & 223MB 	 & 53.17 		  & 82.89 \\
GoogleLeNet 							&	--	     	 &  44MB 	 & 53.63		      & 83.88 \\
VGG-16		 						&	--	     	 & 518MB 	 & 55.24 		  & 84.91 \\
ResNet-152 							& 1 $\times$ 64d	 & 226MB 	 & 54.74		      & 85.08 \\
\hline
ResNeXt-101 (ours)					& 32 $\times$ 4d	 & 165MB 	 & 56.21 	      & 86.25 \\
CRU-Net-116 @$\times$28$\times$14	& 32 $\times$ 4d	 & 163MB 	 & \textbf{56.60} & \textbf{86.55} \\
\hline
\end{tabular}
\end{small}
}
\vskip -0.2in
\end{table}

\section{Conclusion} 
In this work, we introduced a generalized block term decomposition method and revealed its relation with various popular residual functions. Then we unified different residual functions under a single framework and further proposed a novel network architecture called CRU based on the collective tensor factorization. The CRU enables knowledge sharing across different residual units and thus enhances parameter efficiency of the residual units significantly. Employing CRUs achieved the state-of-the-art performance on two large scale benchmark datasets, showing that sharing knowledge throughout the convolutional neural network is  promising  to increase parameter efficiency.

\bibliography{example_paper}
\bibliographystyle{icml2017}

\end{document}